\documentclass{article}
\usepackage{iclr2027_conference,times}

\usepackage{amsmath,amsfonts,bm}

\def\eqref#1{equation~\ref{#1}}

\def\1{\bm{1}}

\DeclareMathAlphabet{\mathsfit}{\encodingdefault}{\sfdefault}{m}{sl}
\SetMathAlphabet{\mathsfit}{bold}{\encodingdefault}{\sfdefault}{bx}{n}

\usepackage{graphicx}
\usepackage{booktabs}
\usepackage{multirow}
\usepackage{amsmath}
\usepackage{makecell}
\usepackage{tabularx}
\usepackage{bm}
\usepackage{float}
\usepackage{xspace}
\usepackage{xcolor}
\usepackage{microtype}
\usepackage{hyperref}
\usepackage{url}
\usepackage{algorithm}
\usepackage{algpseudocode}

\newcommand{\ourname}{DistScene\xspace}

\title{DistScene: Object-to-Scene Distillation for 3D Scene Generation}

\author{\textbf{Kunming Luo}$^{1,2,}$\thanks{Equal contribution. $^\dagger$Correspondence author. Work done during an internship at TeleAI}\quad
\textbf{Hongyu Yan}$^{1,*}$ \quad
\textbf{Ken Deng}$^{1,*}$ \quad
\textbf{Chengcheng Zhou}$^{2}$ \\ 
\textbf{Tianyu Liu}$^{1}$ \quad 
\textbf{Haipeng Li}$^{1}$ \quad
\textbf{Haibin Huang}$^{2}$ \quad
\textbf{Xuelong Li}$^{2,\dagger}$ \quad
\textbf{Ping Tan}$^{1,\dagger}$ \\
$^{1}$The Hong Kong University of Science and Technology
\quad
$^{2}$TeleAI, China Telecom
}
\iclrfinalcopy 
\begin{document}

\maketitle
\lhead{Preprint} %
\begin{figure}[H]
  \centering
  \includegraphics[width=\linewidth,pagebox=cropbox]{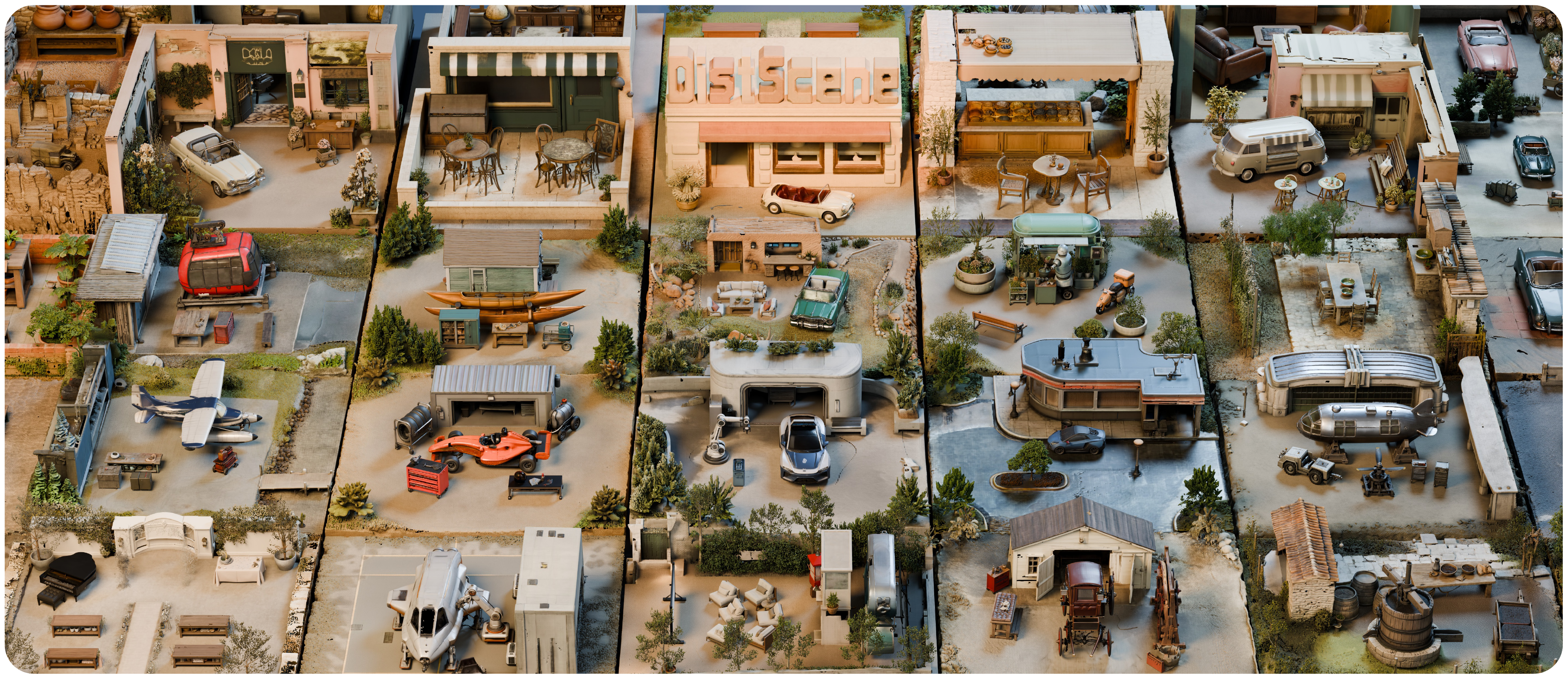}
  \caption{Diverse 3D scenes generated by DistScene from images.}
  \label{fig:teaser}
\end{figure}

\begin{abstract}
We present \ourname{}, a framework for single-image compositional 3D scene generation by jointly modeling the environment and individual objects. Unlike existing methods that represent scenes primarily as collections of objects, we model the environment as an explicit scene component to provide geometric context for object placement. Specifically, we introduce Scene-Frame Generation, which jointly generates separate environment and object components in a shared coordinate frame, allowing their geometry and relative placement to be learned together. Then we introduce Object-Centric Refinement to refine each object in a local frame with scene context. Finally, we develop Object-to-Scene Distillation to transfer pretrained object-generation priors to scene generation through automatically
composed and rendered synthetic scenes. Evaluations on indoor and outdoor benchmarks demonstrate improved scene-level
spatial coherence over the evaluated baselines. Project page: \url{https://coolbeam.github.io/DistScene/}
\end{abstract}

\section{Introduction}


Compositional 3D scene generation is essential for turning images into structured 3D worlds by recovering environments and independent objects, enabling applications in autonomous driving~\cite{lu2025infinicube,liu2026autoweather4d}, virtual reality~\cite{martinez2020unrealrox}, robotics~\cite{wang2024grutopia}, and simulation~\cite{wu2024metaurban}.
Yet this task is challenging because object decomposition, spatial placement,
and 3D geometry generation must be resolved jointly from a single image. 


Existing approaches~\cite{sam3d, 3dfixer} mainly adopt a modular and cascaded paradigm that address these
problems in separate stages, using segmentation~\cite{sam3} or detection~\cite{boxer} for object
decomposition, depth~\cite{moge} or layout estimation~\cite{vggt} for spatial placement, and
object-level 3D generators~\cite{trellis2, hunyuan3d} for geometry recovery.
By leveraging specialized foundation models for these individual
capabilities, such pipelines have substantially improved the generalization
and quality of compositional scene reconstruction.
However, this cascaded paradigm often propagates errors across different intermediate
predictions, leading to incomplete object geometry and inaccurate
spatial relationships. 

To reduce this cascading dependency, recent feed-forward methods~\cite{huang2025midi, lin2026partcrafter} generate
multiple 3D instances and their spatial arrangements within a single learned process.
Although this formulation reduces the need for explicit stage-wise
processing, existing feed-forward methods typically represent a scene as a
collection of object instances without explicitly modeling the surrounding
environment.
Consequently, object arrangements are formed without explicit
constraints from the environmental geometry,
making it difficult to produce globally coherent scene layouts.


Based on these observations, we present DistScene, which directly generates a decomposable 3D scene from a single image by jointly generating the environment and multiple independent objects.
Specifically, we first design a \textbf{Scene-Frame Generation} stage to represent the environment and objects as separate components and jointly generates them in a shared scene frame, preserving component decomposability and scene-level spatial coherence. 
In this way, the environment component can provide a shared spatial context for organizing object placement, relative scales and relationships.

However, this formulation introduces two challenges: objects often occupy a limited portion of the whole scene space and may lose fine geometric details; 
training such a scene generator requires large-scale 3D scenes with decomposed environment and object supervision.
To address the first challenge, we introduce \textbf{Object-Centric
Refinement}, which refines each object in a normalized local space conditioned on the generated scene context. The refined object is then placed back into its original scene location, thereby improving local geometric detail while preserving the scene layout and component relationships. To address the second challenge, we introduce \textbf{Object-to-Scene Distillation}, which procedurally assembles environments and objects generated by a pretrained object generator into 3D scenes. This process transfers object-level generative priors into scalable, component-aligned supervision for training scene generation. 
Extensive experiments across indoor and outdoor settings show that DistScene
generates more coherent scenes than existing approaches while preserving
fine-grained local object details.
Our main contributions are:
\begin{itemize}
    \item We present \textbf{DistScene}, a compositional 3D scene generation framework that jointly generates the environment and independent objects in the scene from a single image.

    \item We introduce \textbf{Object-Centric Refinement} to recover
    fine-grained object geometry in local spaces while preserving the scene layout.

    \item We propose \textbf{Object-to-Scene Distillation} to train our scene generator by automatically constructing scene supervision from a pretrained object generator.
\end{itemize}


\section{Related Work}
\paragraph{Compositional 3D Scene Generation.}
Single-image compositional 3D scene generation~\citep{total3dunderstanding,instpifu,panorecon} aims to recover multiple independent 3D components while preserving their spatial organization. 
Existing methods mainly follow two directions. Component-aligned reconstruction methods~\cite{tang2025zeroscene,zhou2024zero,gen3dsr}  recover individual objects from an image and construct a compositional scene, as exemplified by CAST~\citep{yao2025cast} and SAM3D~\cite{chen2026sam}. These methods typically combine scene decomposition~\cite{sam3,groundedsam,geosam2,sam2}, spatial placement~\cite{mao2025spatiallm,boxer} or object pose estimation~\citep{gen3dsr,scenemaker}, and monocular or multi-view
geometry estimation~\cite{moge,dust3r,vggt,moge2,vggtomga,da3,da2,da,marigold,pi3,d4rt,ppd} to reconstruct the scene
components, providing a reconstruction-based paradigm for
compositional 3D scene generation.
Feed-forward scene generation methods~\citep{huang2025midi,meng2026scenegen} instead generate multiple 3D instances and their spatial arrangements within a single learned process. MIDI~\cite{huang2025midi} and SceneGen~\cite{meng2026scenegen} follow this direction and reduce the need for explicit stage-wise scene construction.
However, these methods remain primarily object-centric and do not explicitly model the scene environment as a generated component. Consequently, the spatial organization of objects is not directly grounded in the geometry of the surrounding environment. In contrast, DistScene jointly generates the environment and multiple objects in a shared scene frame, and further refines each object in a local space while preserving its scene-level
placement.

\paragraph{3D Object Generation.}
Recent image-to-3D methods have established strong generative priors for recovering object geometry and appearance from a single image. Methods based on Vecset representation~\citep{lai2025hunyuan3d,craftman,triposg,detailgen3d,posemaster,clay,hi3dgen,CTR3D}, and Structured 3D Latents~\citep{trellis,trellis2,direct3ds2,sparseflex,sparc3d} improve the quality and diversity of generated 3D assets. Part-aware methods such as PartCrafter~\citep{lin2026partcrafter} further represent an object as a set of structured components, enabling decomposable object generation. Despite these advances, existing object generators are primarily designed for individual objects in a canonical space. Their latent representations describe the geometry and appearance of each object but do not directly
encode its position, scale, or spatial relationship within a surrounding environment. DistScene builds on these object-level priors and extends them
to scene-frame generation, where the environment and individual objects are represented as separate components in a shared scene space.

\section{Preliminary: Sparse-Voxel-Based 3D Generation}
\label{sec:prelim}

Sparse-voxel-based 3D generation represents an asset as active voxels on a regular grid, where each active voxel carries geometry and material features. A sparse VAE, typically built with sparse convolutional networks, compresses this representation into a compact latent space. On top of the latent space, a flow-matching generative model, typically instantiated as a sparse DiT, learns to synthesize assets conditioned on an image. Generation usually proceeds in stages: first predicting the sparse occupancy layout, then generating geometry latents within active voxels, and finally synthesizing appearance latents aligned with the geometry. Formally, this single-asset paradigm can be summarized as
\begin{equation}
\pmb{f} = \{(f_i^{\mathrm{shape}}, f_i^{\mathrm{mat}}, p_i)\}_{i=1}^{L}, \quad
z = \mathcal{E}(\pmb{f}), \quad
\hat{z} \sim \mathcal{G}(\cdot \mid I), \quad
\hat{\pmb{f}} = \mathcal{D}(\hat{z}),
\end{equation}
where \(\mathcal{E}\) and \(\mathcal{D}\) denote the sparse VAE encoder and decoder, \(\mathcal{G}\) is the image-conditioned flow-matching generator, and \(I\) is the input image. Our method follows this paradigm and extends it from a single asset to a complete compositional scene.


\begin{figure}[t]
  \centering
  \includegraphics[width=0.95\linewidth]{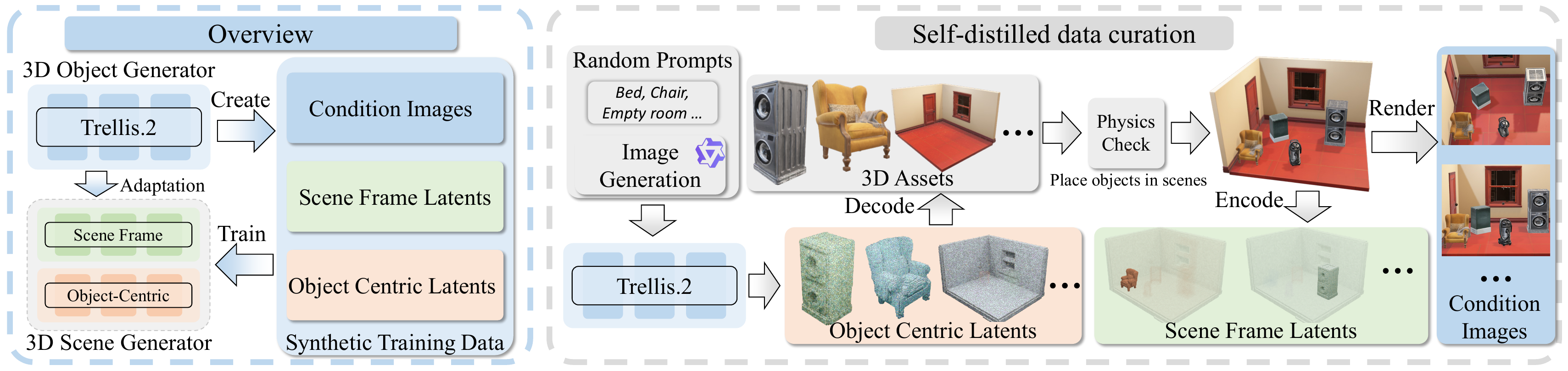}
  \caption{Overview of \textit{Object-to-Scene Distillation}. We adopt TRELLIS.2 as the object generator from which training data are distilled. First, an LLM generates descriptions of 3D assets, including objects and environments; these descriptions are then used to generate the corresponding input images for TRELLIS.2. After the 3D assets are obtained, we place them into the generated environments with physics checks. Finally, we render condition images for each scene as conditioning inputs.}
  \label{fig:data_curation}
\end{figure}

\section{Method}

Given a single image \(I\), our goal is to generate a complete, decomposable compositional scene. We represent the scene in a unified sparse voxel space and jointly synthesize its objects and environment, followed by a scene-conditioned refinement stage that boosts the fidelity of each object. To overcome the scarcity of scene data, we further build a self-distilled data engine that produces large scale physically plausible training scenes. Formally, the overall pipeline can be summarized as
\begin{equation}
S' = \mathcal{R}\big(\mathcal{G}(I)\big), \quad
S = \mathcal{G}(I) = \{O_1,\dots,O_N,E\},
\end{equation}
where \(\mathcal{G}\) jointly generates the objects and the environment from \(I\), and \(\mathcal{R}\) refines each object to produce the final scene \(S'\).

\subsection{Object-to-Scene Distillation}
\label{sec:data_engine}

Training a compositional scene generator requires decomposable scene data with complete meshes for every object. However, existing scene datasets such as 3D-FUTURE~\cite{3dfuture} are limited in both scale and object diversity, which severely bottlenecks the generalization of scene generation models. To this end, we propose Object-to-Scene Distillation to efficiently synthesize scalable training scenes without relying on any existing scene dataset or human annotation.

As illustrated in Fig.~\ref{fig:data_curation}, the core of our Object-to-Scene Distillation is the self-distilled data engine, which generates a scene in three steps. First, an LLM produces a structured description of the scene, specifying the objects it contains and their approximate sizes. Second, an object generator synthesizes each object individually as a complete mesh. Crucially, the same generator also produces the environment, so the background is not an empty space but a geometrically detailed asset, eliminating the need for any separate background model. Third, the generated objects are randomly placed into the generated environment, and the layout is iteratively adjusted by checking their contacts with the environment and collisions with each other. Formally, a scene \(S = \{(O_i, T_i)\}_{i=1}^{N} \cup \{E\}\) is produced by
\begin{equation}
E = \mathcal{G}(d_E), \quad
\{O_i\}_{i=1}^{N} = \{\mathcal{G}(d_i)\}_{i=1}^{N}, \quad
\{T_i\}_{i=1}^{N} = \mathcal{A}\big(\{O_i\}, E\big), \quad
\text{s.t.} \quad \mathcal{C}(S) = 1,
\end{equation}
where \(d_i\) is the LLM-generated description of the \(i\)-th object, \(\mathcal{G}\) is the object generator, \(\mathcal{A}\) is the layout adjustment procedure, \(T_i\) denotes the pose of \(O_i\), and \(\mathcal{C}\) enforces physical plausibility, i.e., no inter-penetration and plausible contacts with the environment. For each generated scene, we render a large number of views and retain only those in which all objects are visible as conditioning images, so that the model learns a consistent image-to-scene mapping. By repeating this process, the engine can produce a scalable source of scenes with diverse objects and layouts, which in turn supervise the scene generator through Object-to-Scene Distillation.

\begin{figure}[t]
  \centering
  \includegraphics[width=0.95\linewidth]{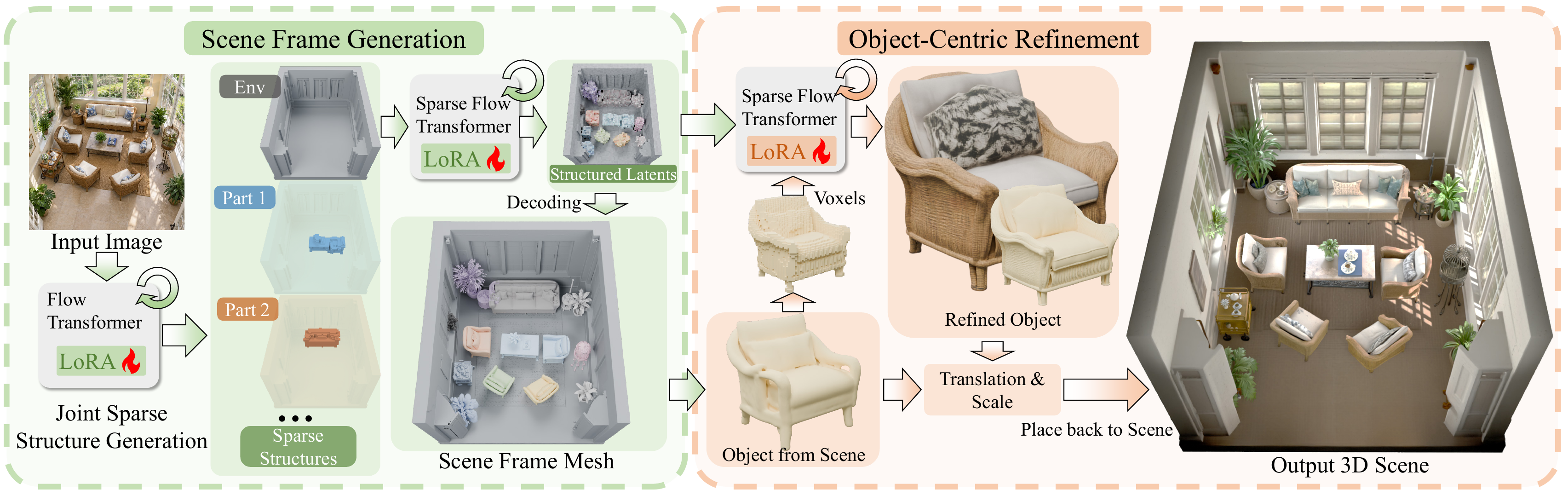}
  \caption{Overview of \textit{Inference Pipeline}: Left: the environment and objects are jointly generated in a shared scene frame through two stages, sparse structure generation and geometry-latent generation. Right: each generated object is transformed to a local voxel support for scene-conditioned refinement and then placed back into the scene using the inverse transformation. }
  \label{fig:inference}
\end{figure}

\subsection{Scene-Frame Generation}
\label{sec:scene_gen}



Existing approaches mainly adopt a modular and cascaded paradigm that address object composition, spatial placement and 3D generation in separate stages. However, this cascaded paradigm often propagates errors across different intermediate predictions. DistScene instead directly generates the environment and multiple independent objects together in a shared scene frame, preserving their component identities and spatial relationships without requiring external segmentation or reconstructed scene geometry at inference.

For representation, we adopt sparse voxels as the unified substrate for both objects and the environment. Given an input image \(I\), we initialize \(N+1\) noise latents, corresponding to \(N\) objects and one environment, denoted as \(\{z_i\}_{i=1}^{N+1}\) with \(z_{N+1}\) reserved for the environment. Crucially, the environment latent acts as a bridge that anchors the positions of all objects and encodes their spatial interactions, thereby coupling the \(N\) object latents into a coherent scene. To jointly denoise these latents, we flatten each latent into a token sequence and concatenate the \(N+1\) sequences along the token dimension into a single sequence \(Z = [z_1; \dots; z_{N+1}]\). Learnable type embeddings are added to distinguish object tokens and environment tokens, and the sparse DiT performs self-attention over \(Z\), allowing all objects and the environment to exchange information during denoising. The generator \(\mathcal{G}\) then jointly denoises \(Z\) conditioned on \(I\) following the same three-stage scheme as in Sec.~\ref{sec:prelim}. Formally,
\begin{equation}
Z = [z_1; \dots; z_{N+1}], \quad
\{x_i\}_{i=1}^{N+1} = \mathcal{G}(Z; I), \quad
S = \mathcal{D}\big(\{x_i\}_{i=1}^{N+1}\big) = \{O_1,\dots,O_N,E\},
\end{equation}
where \(\{x_i\}\) are the generated latents, \(\mathcal{D}\) is the sparse-voxel decoder, and \(S\) is the complete compositional scene.

\begin{figure}[t]
  \centering
  \includegraphics[width=0.95\linewidth,pagebox=cropbox]{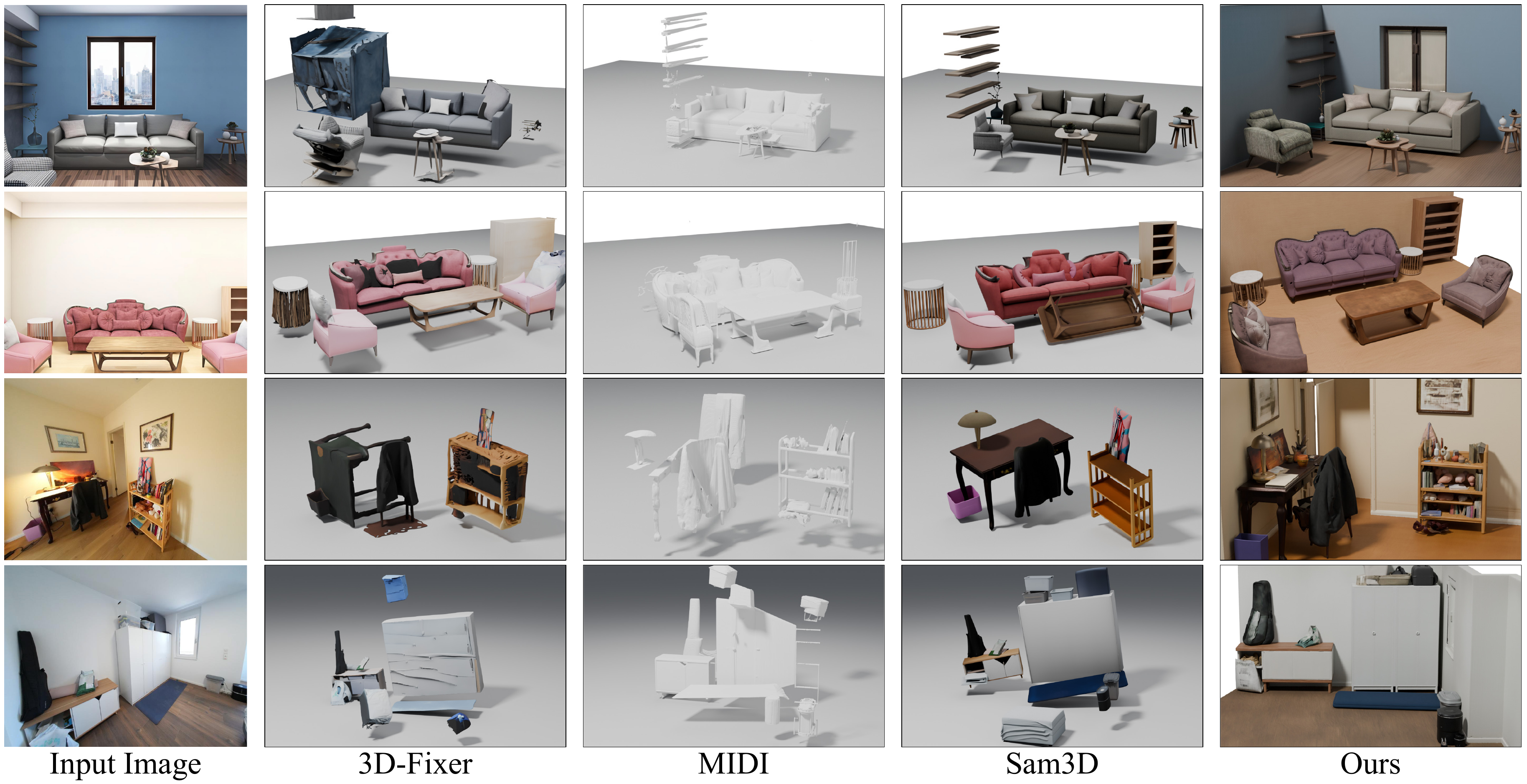}
  \caption{Qualitative comparison with 3D-Fixer, MIDI, and SAM3D on Scannet.}
  \label{fig:gen3dsr_qualitative}
\end{figure}

\subsection{Scene-Aware Object-Centric Refinement}
\label{sec:refine}

After the scene is decomposed into \(N\) objects and an environment, each object occupies only a small fraction of the global sparse voxel grid. Consequently, the effective spatial resolution allocated to each object is far below the maximum that the representation can offer, which limits the fidelity of individual objects. To recover fine geometric details for each component, we introduce a dedicated object refinement stage.

Given the decomposed scene \(\{O_i\}_{i=1}^N \cup \{E\}\), we extract each object \(O_i\), re-center and rescale it into a canonical space, and feed it into a sparse-voxel-based object refinement model \(\mathcal{R}\). To make the refinement aware of the surrounding context, \(\mathcal{R}\) is conditioned jointly on the input image \(I\) and the scene latents produced by \(\mathcal{G}\). Specifically, we annotate the scene latents with a learnable embedding that marks the region corresponding to \(O_i\), so that the refinement process knows which part of the scene information to interact with. The model then refines the object at a higher effective resolution, producing
\begin{equation}
\hat{O}_i = \mathcal{R}\big(O_i \mid I, \{x_j\}_{j=1}^{N+1}, e_i\big),
\end{equation}
where \(\{x_j\}\) are the scene latents and \(e_i\) is the embedding marking \(O_i\)'s corresponding region. The refined objects are then placed back to their original positions, yielding the final scene
\begin{equation}
S' = \{\hat{O}_1,\dots,\hat{O}_N,E\}.
\end{equation}
Since the refinement is conditioned on the scene latents and the marker embedding, the refined object helps preserve spatial and semantic correspondence with the scene when placed back, while still benefiting from the higher effective resolution. This refinement stage operates entirely in the same sparse voxel domain and can be seamlessly integrated with the generator, allowing each object to be refined independently without affecting the global scene layout.

\subsection{Training Objective and Network Architecture}
\label{sec:training}

We instantiate the scene generator \(\mathcal{G}\) as a sparse DiT and fine-tune it with LoRA, since the underlying sparse-voxel generator already possesses strong image-to-3D priors and only needs to adapt to the compositional setting. Training follows the standard flow-matching objective used in sparse-voxel generation, applied separately to each object latent and to the environment latent:
\begin{equation}
\mathcal{L}_{\text{scene}} = \sum_{i=1}^{N+1} \mathbb{E}_{t, z_0^i, z_1^i} \left\| v_\theta(z_t^i, t) - (z_1^i - z_0^i) \right\|^2,
\end{equation}
where \(i=1,\dots,N\) indexes the \(N\) objects and \(i=N+1\) indexes the environment, \(z_0^i\) and \(z_1^i\) are the noise and data latents, and \(v_\theta\) is the predicted velocity. This loss jointly supervises the flow of every object and the scene, while LoRA adapters keep the pretrained backbone frozen, enabling efficient fine-tuning while retaining the pretrained generative prior.

For the object refinement model \(\mathcal{R}\), we construct coarse-to-fine training pairs without any manual annotation. Given a high-quality object latent \(z_1\), we downsample it, pass it through the sparse VAE encoder and decoder, and upsample it back to obtain a degraded coarse latent \(z_0\). The refinement model is trained to recover \(z_1\) from \(z_0\) with the same flow-matching objective:
\begin{equation}
\mathcal{L}_{\text{refine}} = \mathbb{E}_{t, z_0, z_1} \left\| v_\phi(z_t, t \mid I,\{x_j\}_{j=1}^{N+1}, e) - (z_1 - z_0) \right\|^2,
\end{equation}
where \(v_\phi\) is the refinement velocity conditioned on the input image \(I\) and the marker embedding \(e\) that indicates the corresponding scene region. This allows the refinement module to learn detail recovery entirely from self-generated pairs, without requiring any external high-resolution data.

\begin{figure}[t]
  \centering
  \includegraphics[width=0.95\linewidth,pagebox=cropbox]{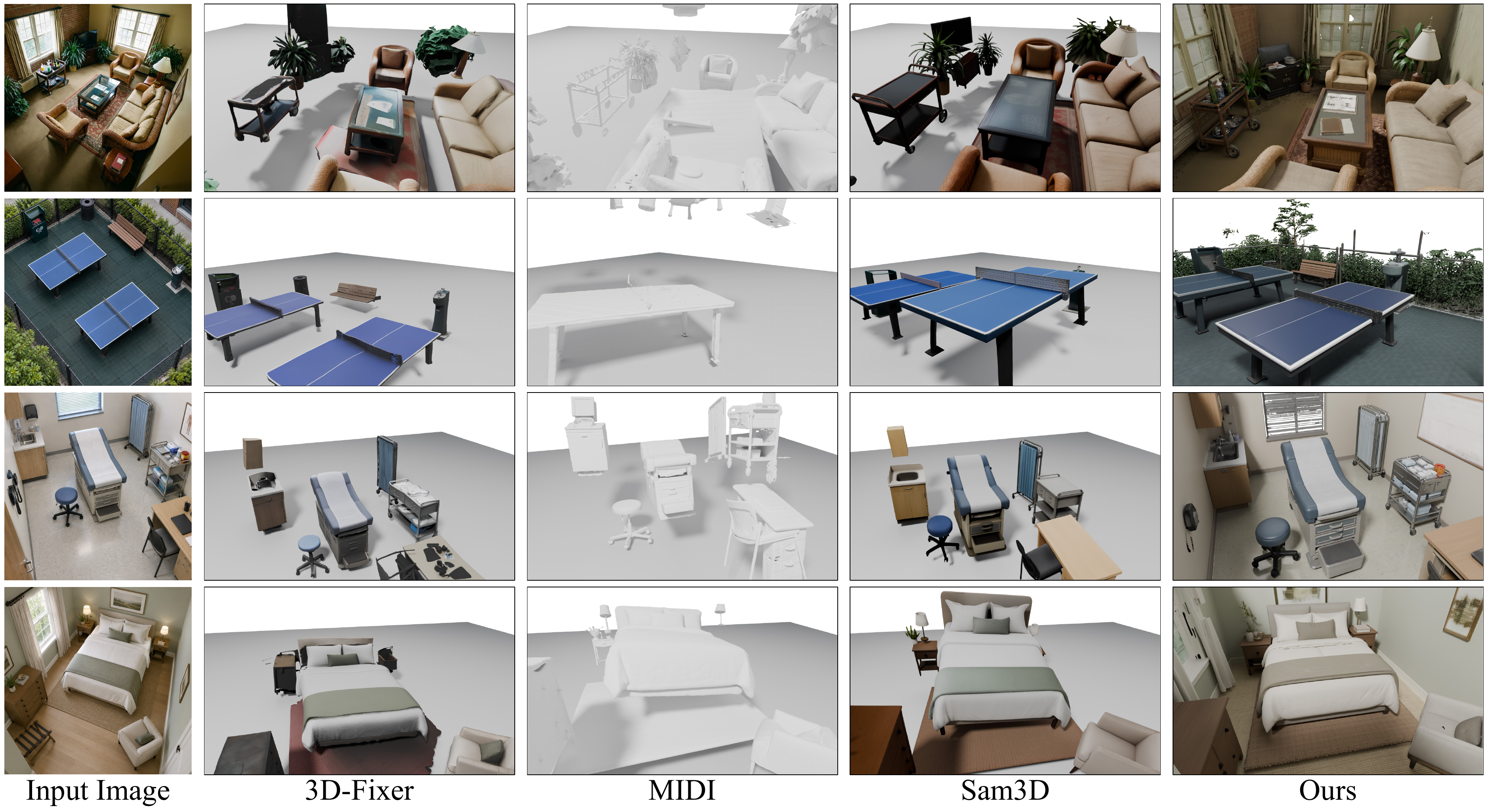}
  \caption{Qualitative comparison on text-to-image-generated inputs beyond the benchmarks.}
  \label{fig:free_style_qualitative}
\end{figure}

\section{Experiments}
\label{sec:experiments}
\subsection{Implementation Details}
We use TRELLIS.2 as the pretrained object generator and adapt it with LoRA with a rank of 32. For Object-to-Scene Distillation, we construct about $125k$ synthetic training scenes from $70k$ generated objects and $46k$ empty scenes, including $106k$ indoor scenes and $18k$ outdoor scenes. 
All synthetic training data are generated at a resolution of 512. Although the model is trained only at this resolution, we find that the adapted model supports inference at a resolution of 1024. Unless otherwise specified, all main experiments use 1024-resolution Scene-Frame Generation and Object-Centric Refinement. For evaluation, we follow 3D-Fixer to evaluate indoor scene generation on MIDI-test and Gen3DSR-test, reporting scene-level CD and F-score. On MIDI-test, we also report object-level CD and F-score, together with bounding-box IoU in the scene coordinate system. For outdoor scenes, we
follow the settings of Extend3D and evaluate on UrbanScene3D, reporting CD-L1 and F-score.

\subsection{Main Comparison}
\paragraph{Outdoor scene generation.}
Table~\ref{tab:urbanscene_geometry} compares outdoor scene geometry on UrbanScene3D. DistScene achieves the lowest CD-L1 (0.0772) and highest F-score (0.722) among the evaluated methods. Compared with Extend3D, CD-L1 decreases from 0.0832 to 0.0772, while F-score increases from 0.680 to 0.722. DistScene also improves both metrics over TRELLIS.2, showing the effectiveness of our object-scene distillation for outdoor scene generation.

\paragraph{Indoor scene generation.}
Table~\ref{tab:indoor_geometry} reports results on MIDI-test and
Gen3DSR-test. DistScene achieves the best scene-level CD and F-score on both benchmarks. On MIDI-test, it reduces scene CD from 0.1295 to 0.0877, a relative improvement of 32.2\% over 3D-Fixer, and increases scene F-score from 65.08 to 71.59, an improvement of 6.51 percentage points. It also achieves the highest bounding-box IoU, as well as the best object-level CD and F-score among the compared methods. On Gen3DSR-test, DistScene obtains a scene CD of 0.0958 and an F-score of 79.68\%, compared with 0.1027 and 77.97\% for 3D-Fixer. These results demonstrate that DistScene improves both
scene-level spatial alignment and object-level geometry on indoor scene generation.

\begin{table*}[t]
  \centering
  \normalsize
  \setlength{\tabcolsep}{3pt}
  \renewcommand{\arraystretch}{1.05}
  \setlength{\extrarowheight}{1pt}
  \resizebox{0.80\linewidth}{!}{
  \begin{tabular}{
    >{\arraybackslash}p{1.4cm}| 
    >{\centering\arraybackslash}p{2.2cm} 
    >{\centering\arraybackslash}p{1.4cm} 
    >{\centering\arraybackslash}p{1.4cm} 
    >{\centering\arraybackslash}p{1.4cm} 
    >{\centering\arraybackslash}p{1.4cm} 
    >{\centering\arraybackslash}p{1.4cm} 
  }
    \hline
    Metric
    & Hunyuan3D 2.1
    & TRELLIS
    & EvoScene
    & Extend3D
    & TRELLIS.2
    & ours \\
    \hline
    CD-L1 $\downarrow$
    & 0.1933
    & 0.1397
    & 0.1120
    & 0.0832
    & 0.0831
    & \textbf{0.0772} \\
    F@0.05 $\uparrow$
    & 0.411
    & 0.556
    & 0.543
    & 0.680
    & 0.701
    & \textbf{0.722} \\
    \hline
  \end{tabular}}
  \caption{Geometry comparison on the outdoor UrbanScene3D benchmark. Lower CD-L1 and higher F-score indicate better reconstruction.}
  \label{tab:urbanscene_geometry}
\end{table*}

\begin{table}[t]
  \centering
  \normalsize
  \setlength{\tabcolsep}{3pt}
  \renewcommand{\arraystretch}{1.05}
  \resizebox{0.90\linewidth}{!}{
    \begin{tabular}{
      >{\arraybackslash}p{2.7cm}| 
      >{\centering\arraybackslash}p{1.25cm} 
      >{\centering\arraybackslash}p{1.25cm} 
      >{\centering\arraybackslash}p{1.25cm} 
      >{\centering\arraybackslash}p{1.25cm} 
      >{\centering\arraybackslash}p{1.25cm}| 
      >{\centering\arraybackslash}p{1.25cm} 
      >{\centering\arraybackslash}p{1.25cm} 
    }
    \hline
    \multirow{2}{*}{Methods} & \multicolumn{5}{c|}{MIDI-test} & \multicolumn{2}{c}{Gen3DSR-test} \\
    \cline{2-8}
    & CD$_\mathrm{S}$ $\downarrow$ & FS$_\mathrm{S}$ $\uparrow$
    & CD$_\mathrm{O}$ $\downarrow$ & FS$_\mathrm{O}$ $\uparrow$ & IoU $\uparrow$
    & CD$_\mathrm{S}$ $\downarrow$ & FS$_\mathrm{S}$ $\uparrow$ \\
    \hline
    Gen3DSR & 0.3795 & 26.45 & 0.2609 & 28.73 & 0.0712 & 0.6701 & 11.33 \\
    TRELLIS.2 & 0.1444 & 54.19 & 0.2475 & 31.07 & 0.1471 & 0.2099 & 44.44 \\
    MIDI & 0.1550 & 53.46 & 0.2162 & 36.26 & 0.1639 & 0.2830 & 32.53 \\
    SAM3D & 0.1486 & 59.79 & 0.1607 & 49.72 & 0.2102 & 0.2055 & 48.73 \\
    SceneGen & 0.1502 & 51.40 & 0.1904 & 40.88 & 0.1217 & 0.2040 & 44.31 \\
    3D-Fixer & 0.1295 & 65.08 & 0.1704 & 48.10 & 0.3527 & 0.1027 & 77.97 \\
    \hline
    DistScene (ours) & \textbf{0.0877} & \textbf{71.59} & \textbf{0.1529} & \textbf{51.64} & \textbf{0.3688} & \textbf{0.0958} & \textbf{79.68} \\
    \hline
    \end{tabular}}
  \caption{Geometry comparison on MIDI-test and Gen3DSR-test. S/O denote scene/object level; IoU measures bounding-box overlap. F-scores use a 0.1 threshold and are reported in percent.}
  \label{tab:indoor_geometry}
\end{table}

\paragraph{Qualitative comparison.}
Figure~\ref{fig:gen3dsr_qualitative} presents scene generation results using test images from Gen3DSR and real images from ScanNet. Compared with the input images, 3D-Fixer, MIDI, and SAM3D can recover the main objects, but often produce incomplete geometry or misaligned spatial relationships between objects. In contrast, DistScene jointly recovers the environment and multiple independent objects, while better preserving the relative layout and support relationships in the input image. These observations are consistent with the quantitative results in Table~\ref{tab:indoor_geometry}.

Figure~\ref{fig:free_style_qualitative} further presents examples using text-to-image-generated inputs. Despite the greater diversity in viewpoints, scene types, and object compositions, DistScene consistently generates environments and objects corresponding to the input content while preserving reasonable scene structures, object layouts, and component relationships.

\begin{table*}[t]
  \centering
  \small
  \setlength{\tabcolsep}{3pt}
  \renewcommand{\arraystretch}{1.05}
  \resizebox{0.90\linewidth}{!}{
  \begin{tabular}{
    >{\arraybackslash}p{3.2cm}| 
    >{\centering\arraybackslash}p{1.6cm} 
    >{\centering\arraybackslash}p{1.6cm} 
    >{\centering\arraybackslash}p{1.6cm} 
    >{\centering\arraybackslash}p{1.6cm} 
    >{\centering\arraybackslash}p{1.6cm} 
  }
    \hline
    Metric
    & \makecell{TRELLIS.2}
    & 3D-Fixer
    & SAM3D
    & Extend3D
    & Ours \\
    \hline
    Image consistency
    & 8.54
    & 1.71
    & 0.61
    & 27.20
    & \textbf{61.95} \\
    Geometry quality
    & 9.15
    & 3.29
    & 0.61
    & 19.15
    & \textbf{67.80} \\
    Appearance quality
    & 8.54
    & 3.41
    & 0.37
    & 21.22
    & \textbf{66.46} \\
    Average
    & 8.74
    & 2.80
    & 0.53
    & 22.52
    & \textbf{65.41} \\
    \hline
  \end{tabular}}
  \caption{User study preference rates (\%). Higher values indicate stronger
  participant preference.}
  \label{tab:user_study}
\end{table*}

\paragraph{User study.}
To evaluate perceptual quality, we conduct a user study with 43 participants
on 20 randomly selected text-to-image-generated inputs. For each input,
participants view anonymized outputs from five methods in randomized order and
select the best result for image consistency, geometry quality, and appearance
quality. We compute each method's winning probability as its fraction of
selections and report the average across the three criteria. As shown in
Table~\ref{tab:user_study}, DistScene achieves the highest preference rate on
all three criteria and the highest average winning probability.

\begin{table}[t]
  \centering
  \normalsize
  \setlength{\tabcolsep}{3pt}
  \renewcommand{\arraystretch}{1.05}
  \resizebox{0.90\linewidth}{!}{
    \begin{tabular}{
      >{\arraybackslash}p{3.2cm}| 
      >{\centering\arraybackslash}p{1.25cm} 
      >{\centering\arraybackslash}p{1.25cm} 
      >{\centering\arraybackslash}p{1.25cm} 
      >{\centering\arraybackslash}p{1.25cm} 
      >{\centering\arraybackslash}p{1.25cm}| 
      >{\centering\arraybackslash}p{1.25cm} 
      >{\centering\arraybackslash}p{1.25cm} 
    }
    \hline
    \multirow{2}{*}{Method} & \multicolumn{5}{c|}{MIDI-test} & \multicolumn{2}{c}{Gen3DSR-test} \\
    \cline{2-8}
    & CD$_\mathrm{S}$ $\downarrow$ & FS$_\mathrm{S}$ $\uparrow$
    & CD$_\mathrm{O}$ $\downarrow$ & FS$_\mathrm{O}$ $\uparrow$ & IoU $\uparrow$
    & CD$_\mathrm{S}$ $\downarrow$ & FS$_\mathrm{S}$ $\uparrow$ \\
    \hline
    TRELLIS.2 (no mask) & 0.1610 & 53.40 & 0.2222 & 34.46 & 0.1798 & 0.2177 & 41.45 \\
    TRELLIS.2 (mask) & 0.1500 & 55.57 & 0.2280 & 34.12 & 0.1900 & 0.2099 & 44.44 \\
    \hline
    SF w/o Env & 0.0922 & 67.94 & 0.2444 & 29.98 & 0.1798 & 0.1367 & 62.04 \\
    SF w/o Env (MIDI) & 0.0981 & 65.74 & 0.2347 & 33.03 & 0.2084 & 0.1936 & 46.68 \\
    SF w. Env & 0.0928 & 68.86 & 0.2037 & 40.98 & 0.2544 & 0.1096 & 73.21 \\
    SF + Refine w/o Ctx & 0.1055 & 66.54 & 0.1913 & 41.97 & 0.2964 & 0.1169 & 69.82 \\
    SF + Refine (full) & \textbf{0.0877} & \textbf{71.59} & \textbf{0.1529} & \textbf{51.64} & \textbf{0.3688} & \textbf{0.0958} & \textbf{79.68} \\
    \hline
    \end{tabular}}
  \caption{Ablation study on MIDI-test and Gen3DSR-test. SF denotes
Scene-Frame Generation, Env denotes environment generation, Refine denotes
Object-Centric Refinement, and Ctx denotes scene-frame context.}
  \label{tab:ablation_components}
\end{table}
\begin{table}[t]
  \centering
  \normalsize
  \setlength{\tabcolsep}{3pt}
  \renewcommand{\arraystretch}{1.0}
  \resizebox{0.90\linewidth}{!}{
    \begin{tabular}{
      >{\arraybackslash}p{3.6cm}| 
      >{\centering\arraybackslash}p{1.25cm} 
      >{\centering\arraybackslash}p{1.25cm}| 
      >{\centering\arraybackslash}p{1.25cm} 
      >{\centering\arraybackslash}p{1.25cm}| 
      >{\centering\arraybackslash}p{1.25cm} 
      >{\centering\arraybackslash}p{1.25cm}| 
      >{\centering\arraybackslash}p{1.25cm} 
    }
    \hline
    \multirow{2}{*}{Method} & \multicolumn{2}{c|}{MIDI-test}
    & \multicolumn{2}{c|}{Gen3DSR-test}
    & \multicolumn{2}{c|}{Time $\downarrow$}
    & \multirow{2}{*}{\makecell{Memory\\(GB) $\downarrow$}} \\
    \cline{2-7}
    & CD$_\mathrm{S}$ $\downarrow$ & FS$_\mathrm{S}$ $\uparrow$
    & CD$_\mathrm{S}$ $\downarrow$ & FS$_\mathrm{S}$ $\uparrow$
    & \makecell{Scene\\(s)} & \makecell{Refine\\(s/obj)} & \\
    \hline
    SF-512 & 0.1318 & 61.77 & 0.1077 & 71.96 & \textbf{5.7} & -- & \textbf{17.9} \\
    SF-1024 & 0.0928 & 68.86 & 0.1096 & 73.21 & 36.4 & -- & 21.4 \\
    SF-512 + Ref-512 & 0.0934 & 68.35 & 0.0974 & 78.35 & \textbf{5.7} & \textbf{3.3} & 18.0 \\
    SF-1024 + Ref-1024 (full) & \textbf{0.0877} & \textbf{71.59} & \textbf{0.0958} & \textbf{79.68} & 36.4 & 38.5 & 21.8 \\
    \hline
    \end{tabular}}
  \caption{Ablation of resolution for refinement and efficiency.}
  \label{tab:ablation_resolution}
\end{table}

\begin{figure}[t]
  \centering
  \includegraphics[width=0.90\linewidth,pagebox=cropbox]{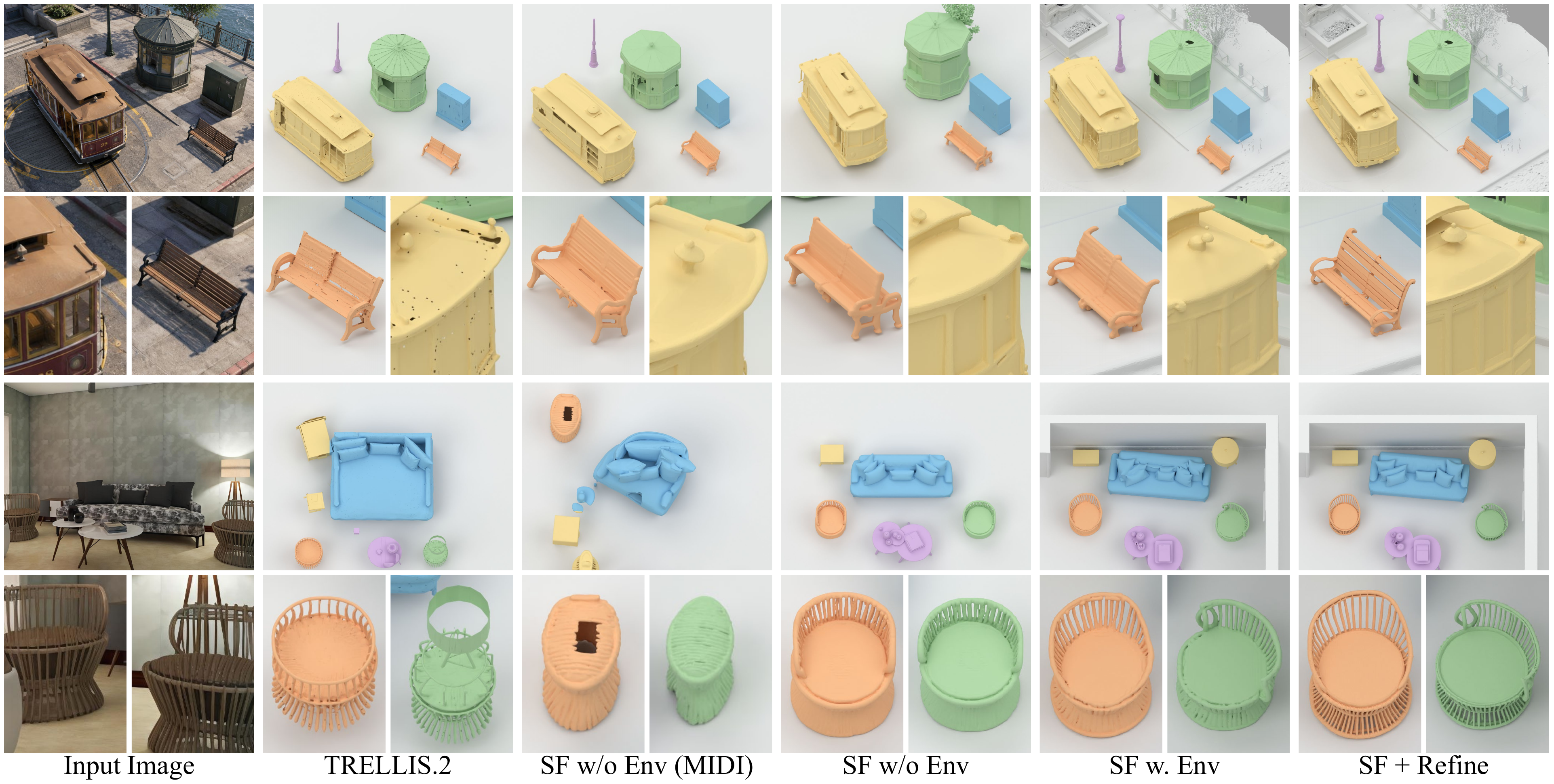}
  \caption{Qualitative ablation study of training supervision, environment generation, and refinement.}
  \label{fig:ablation_visual}
\end{figure}


\subsection{Ablation Studies}
\label{sec:ablations}

We ablate the three key components of DistScene: environment generation in
the shared scene frame, the distilled scene training data, and
scene-conditioned object refinement. We further analyze the effect of
generation resolution and report the runtime and memory cost of each
configuration. In Table~\ref{tab:ablation_components}, SF denotes
Scene-Frame Generation without local refinement, Env denotes the explicit
environment component, and Ctx denotes the scene-frame context used during
refinement. Unless otherwise specified, all variants operate at a resolution
of 1024.

\paragraph{Effect of environment generation.}
We first compare SF w/o Env and SF w. Env to evaluate the effect of jointly
generating the environment with the objects. Adding the environment
substantially improves object-level CD and F-score on MIDI-test, increasing
the object F-score from 29.98 to 40.98, while improving bounding-box IoU from
0.1798 to 0.2544. It also reduces scene CD from 0.1367 to 0.1096 and raises
the F-score from 62.04 to 73.21 on Gen3DSR-test. The MIDI scene CD remains
nearly unchanged, while its scene F-score improves from 67.94 to 68.86.
These results indicate that explicitly generating the environment provides
useful scene-level context for organizing object geometry and spatial
relationships. The qualitative results in
Figure~\ref{fig:ablation_visual} show the same effect: without the
environment component, the chair geometry in the first example is degraded,
and the table lamp in the second indoor scene is missing.

\paragraph{Effect of Object-to-Scene Distillation.}
To isolate the effect of our distilled training data, we train the same
object-only architecture on either our data or the MIDI training set,
corresponding to SF w/o Env and SF w/o Env (MIDI), respectively. Both
MIDI-test and Gen3DSR-test are out of domain for our synthetic training data,
whereas MIDI-test is in domain for the MIDI training set. Despite this
disadvantage, our data improves the scene-level metrics on MIDI-test and
provides a much larger gain on Gen3DSR-test, reducing scene CD from 0.1936 to
0.1367 and increasing F-score from 46.68 to 62.04. The MIDI-trained variant
retains an expected advantage on the object-level metrics and bounding-box
IoU of MIDI-test. Compared with the pretrained TRELLIS.2 baseline supplied
with object masks, SF w/o Env also achieves substantially stronger
scene-level performance on both benchmarks. These comparisons show that our
distilled data transfers the object generator toward scene-level generation
and generalizes beyond its synthetic training distribution.

\paragraph{Effect of Object-Centric Refinement.}
Comparing SF w. Env with the full model shows that Object-Centric Refinement
consistently improves the quality. On MIDI-test, it
reduces object CD from 0.2037 to 0.1529, increases object F-score from 40.98
to 51.64, and improves bounding-box IoU from 0.2544 to 0.3688. On
Gen3DSR-test, scene CD decreases from 0.1096 to 0.0958 and F-score increases
from 73.21 to 79.68. We further remove the scene-frame latent context during
refinement. This variant performs markedly worse than the full model across
all metrics. Without scene context, the refinement model has difficulty
associating the target component with its corresponding object in the input
image, demonstrating that scene-conditioned refinement is essential for
recovering the correct local geometry.

\paragraph{Resolution and efficiency.}
Table~\ref{tab:ablation_resolution} reports the quality, runtime, and peak
memory of different scene-frame and refinement resolutions. Increasing the
resolution consistently improves reconstruction quality at additional
runtime and modest memory cost. The 512-resolution variant with refinement
offers a favorable efficiency--quality trade-off, requiring 5.7 seconds for
scene generation and 3.3 seconds per refined object with 18.0\,GB peak
memory, while the full 1024-resolution model achieves the best accuracy with
21.8\,GB peak memory.
\section{Conclusion}
We introduced \ourname{}, which treats the environment as an explicit
component of compositional 3D scene generation. This shared scene-frame
formulation enables the environment and objects to be generated together,
providing a common geometric context for their spatial arrangement.
Object-Centric Refinement then restores fine object geometry in local
coordinate frames, and Object-to-Scene Distillation supplies the aligned
training supervision required by the framework. Experiments on indoor and outdoor benchmarks show that \ourname{} produces
superior results than existing methods.

\bibliography{iclr2027_conference}
\bibliographystyle{iclr2027_conference}

\appendix
\clearpage
\section{Supplementary Material}

In this supplement, we first provide the network architecture detail in Sec.~\ref{sec:supp_network}. We also provide further
data curation details in Sec.~\ref{sec:supp_data} and further comparison in Sec.~\ref{sec:supp_resolution}. Finally, we provide discussion on limitation and more qualitative results in Sec.~\ref{sec:supp_failure_cases} and Sec.~\ref{sec:supp_more_results}. We encourage the readers to view our accompanying videos in the supplementary materials.

\subsection{Network Architecture}
\label{sec:supp_network}

\begin{figure}[t]
  \centering
  \includegraphics[width=\linewidth,pagebox=cropbox]
  {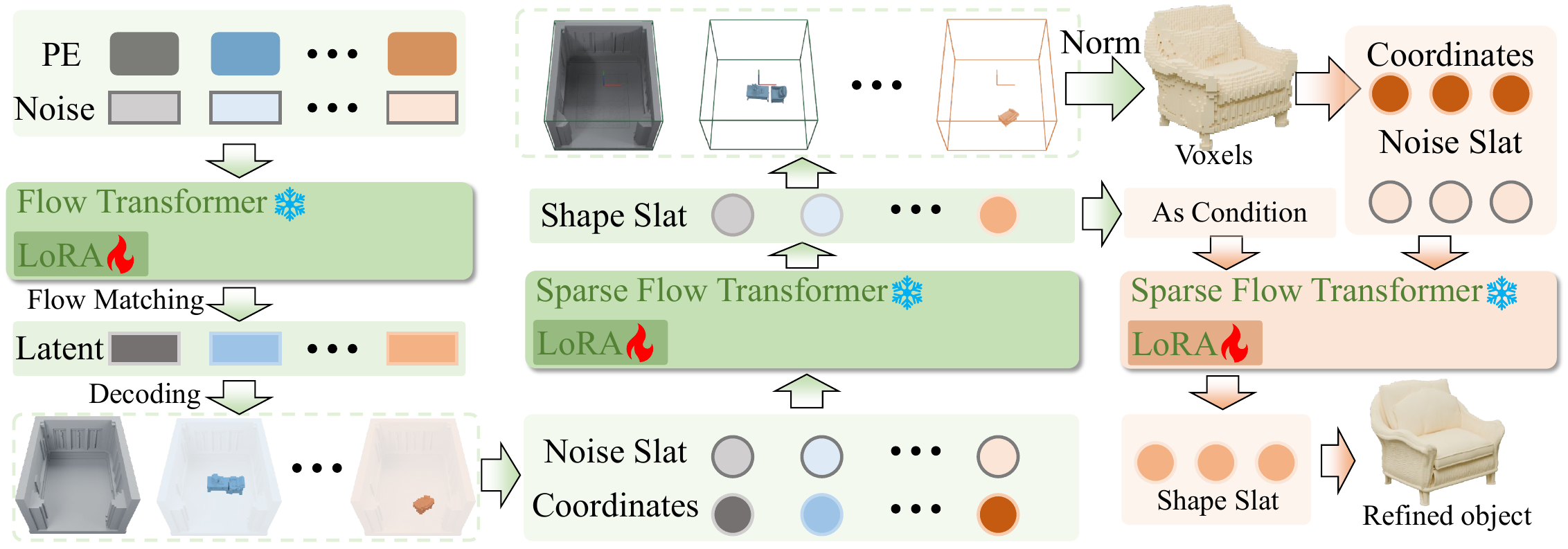}
  \caption{Detailed network architecture of DistScene.
  Left: a flow transformer jointly generates sparse-structure latents for
  the environment and individual objects, which are decoded into
  scene-frame voxel supports.
  Middle: a sparse flow transformer generates component geometry latents
  on these supports.
  Right: the target object is normalized to construct a local voxel
  support, on which a refinement transformer generates local geometry
  conditioned on the fixed scene-frame latents.
  Colors indicate component correspondence; snowflakes and flames denote
  frozen pretrained transformer weights and trainable LoRA adapters.
  Image conditioning and appearance generation are omitted for clarity.}
  \label{fig:supp_network}
\end{figure}

Figure~\ref{fig:supp_network} illustrates the three geometry-generation
modules of DistScene: joint sparse-structure generation, scene-frame
geometry generation, and scene-conditioned object refinement. Colors
indicate corresponding scene components across modules. The snowflake and
flame symbols denote frozen pretrained transformer weights and trainable
LoRA adapters, respectively.

\paragraph{Sparse voxel and latent representations.}
We use sparse voxels as the common spatial substrate for all scene components.
A voxel denotes an active cell on a 3D grid, while a voxel support is the set
of active voxel coordinates occupied by a component. The support describes the
coarse spatial extent of the component and does not itself contain appearance
information. The sparse-structure module predicts a latent representation that
is decoded into these voxel coordinates:
\[
H_i
\ \xrightarrow{\mathcal{D}_{\mathrm{SS}}}\
S_i,
\]
where $H_i$ is the structure latent and $S_i$ is the support of component $i$.

A Structured Latent (SLAT) assigns a learned feature vector to every active
voxel in a support. We use separate shape and texture SLATs. Shape SLATs encode
the geometric features used to decode the component mesh, whereas texture SLATs
encode appearance and material information conditioned on the geometry:
\[
(S_i,Z_i^{\mathrm{shape}})
\ \xrightarrow{\mathcal{D}_{\mathrm{shape}}}\
M_i,
\qquad
(S_i,Z_i^{\mathrm{shape}},Z_i^{\mathrm{tex}})
\ \xrightarrow{\mathcal{D}_{\mathrm{tex}}}\
(M_i,\mathcal{A}_i).
\]
Scene-frame supports and SLATs use the shared scene coordinate system.
Object-centric supports and SLATs use a normalized local coordinate system
defined for an individual object. The component identity is preserved by the
corresponding type or component embedding.

\paragraph{Joint sparse-structure generation.}
We allocate a latent representation to each scene component, including the
environment and the individual objects. Starting from noise, the flow
transformer jointly generates the component structure latents, using learnable type embeddings to distinguish their corresponding slots.
Decoding these latents produces separate sparse voxel supports in the
shared scene frame. Each support specifies the spatial extent and coarse
structure of its component while retaining its identity.

\paragraph{Scene-frame geometry generation.}
The generated voxel supports provide the coordinates for the subsequent
geometry-generation stage. Conditioned on the input image, a sparse flow
transformer jointly generates geometry features on the active voxels of
all components. These features form the structured geometry latents,
denoted as Shape SLat in Figure~\ref{fig:supp_network}. Since the voxel
coordinates are defined in a shared scene frame, the decoded environment
and object meshes can be assembled directly while remaining separate
components.

\paragraph{Scene-conditioned object refinement.}
For each target object, we normalize its scene-frame mesh through
translation and uniform scaling and voxelize the normalized mesh to
construct a local sparse support. This transformation preserves the
object's orientation and increases the effective spatial resolution
available for its geometry. The refinement module generates a new local
geometry latent on this support, conditioned on the input image and the
fixed scene-frame geometry latents of the environment and all objects.
Component correspondence is retained through the learnable component embeddings,
allowing the local target to be associated with its scene-frame component.
During refinement, only the target object's local latent is generated;
the scene-frame latents remain unchanged.

The refined latent is decoded into an object mesh and placed back into
the scene using the recorded scene-space placement and scale. Repeating
this procedure for each object yields a refined compositional scene while
retaining the generated environment. Appearance generation is conditioned
on the corresponding geometry and is omitted from
Figure~\ref{fig:supp_network} for clarity.

\subsection{Training Data Curation}
\label{sec:supp_data}

\paragraph{Asset generation and scene composition.}
We sample text prompts for object assets and empty environments, generate their
conditioning images, and use the pretrained object generator to obtain
canonical shape SLATs, texture SLATs, and decoded meshes. For each environment,
we sample a small set of compatible objects and place them in the environment
coordinate system using sampled scales and translations. The component identity,
canonical latents, and scene-space transform are retained for every object.

The composition is accepted only after geometry-based physical plausibility
checks. We first estimate valid floor regions and support surfaces. Each object
is dropped onto a valid support surface, and placements that float above or
penetrate the support are discarded. Object--object intersections are first
filtered with an axis-aligned bounding-box test. For overlapping pairs, we
further inspect the corresponding mesh regions and adjust the object position
when the overlap is resolvable; scenes with unresolved intersections are
discarded. We additionally check collisions between objects and non-floor
environment surfaces.

For each accepted composition, we render conditioning images from sampled
viewpoints and re-encode the placed environment and objects in the shared
scene frame to obtain scene-frame structure and SLAT targets. The original
canonical shape and texture SLATs produced during asset generation are retained
as the object-centric targets for refinement and texture generation.

\begin{figure}[t]
  \centering
  \includegraphics[width=\linewidth,pagebox=cropbox]
  {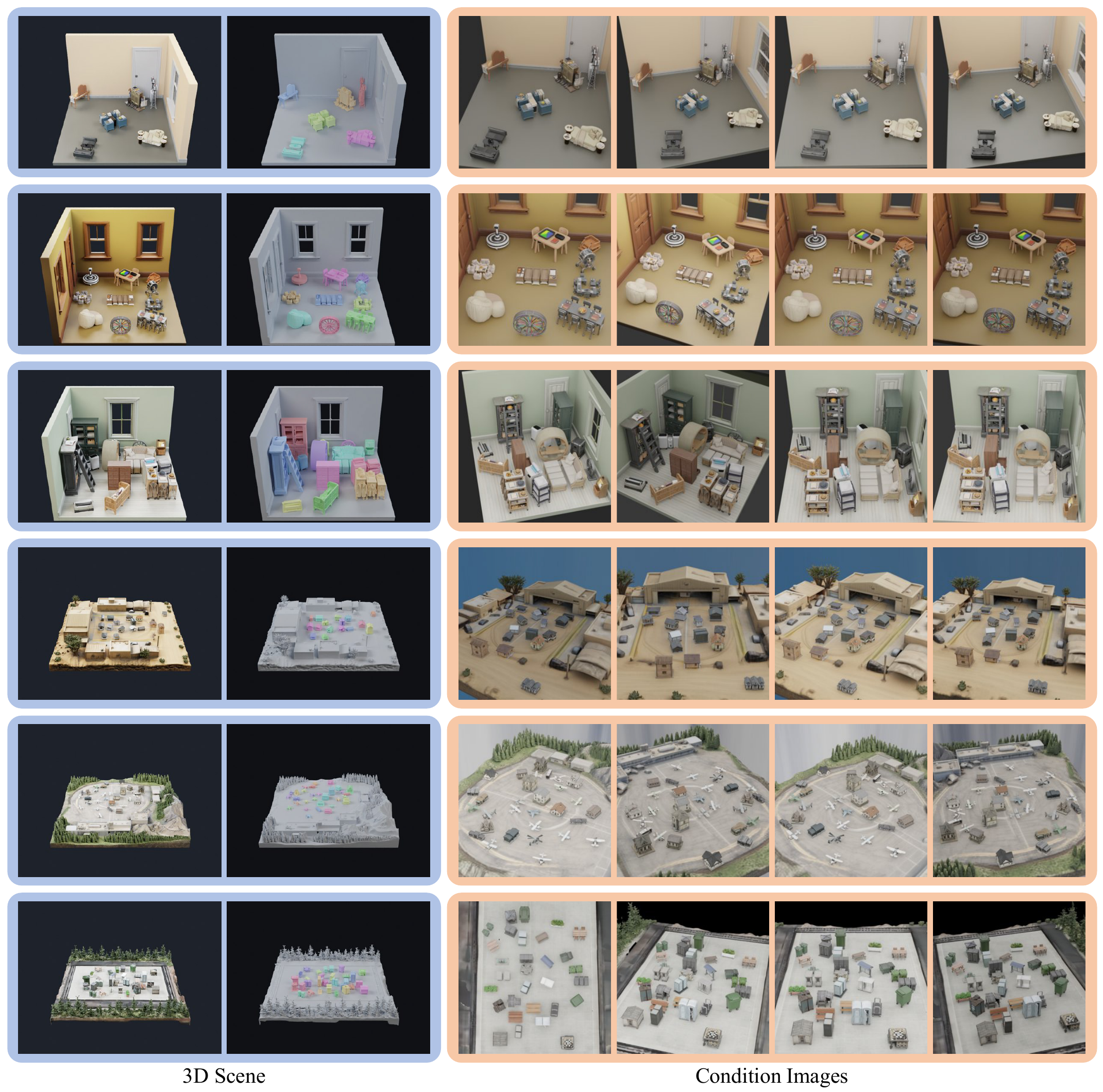}
  \caption{Visualization of randomly sampled synthetic training data generated by our
  Object-to-Scene Distillation pipeline. The left block shows composed 3D
  scenes and their component-level visualizations, while the right block
  shows multi-view conditioning images rendered from the same scenes. The
  examples cover diverse indoor and outdoor environments, object
  compositions, and spatial layouts.}
  \label{fig:our_synthetic_data}
\end{figure}

Figure~\ref{fig:our_synthetic_data} visualizes randomly sampled synthetic
training examples produced by our Object-to-Scene Distillation pipeline. The
left block shows composed 3D scenes and their component-level visualizations,
where the environment and individual objects remain separately identifiable.
The right block shows multiple conditioning images rendered from the
corresponding scenes. The examples include both indoor and outdoor
environments with diverse object compositions and spatial layouts.

\subsection{Training and Inference Details}
\label{sec:supp_implementation}
We use TRELLIS.2 as the pretrained object generator and adapt the trainable
generation modules with LoRA of rank 32, while keeping the pretrained
backbones frozen. We optimize the LoRA parameters with AdamW using a learning
rate of $1\times10^{-4}$ and a global batch size of 64. Each adapted module is
trained for 20,000 iterations on the 512-resolution training targets. At test
time, the same modules are run at 1024 resolution for the main experiments.

Due to the GPU memory limitation, we cap training scenes at 30 components, including the environment, and randomly pad each component sequence by appending a variable number of empty object latents. These empty slots are represented by empty
supports and are included during training, allowing the model to be queried
with a fixed component count larger than the number of visible objects. At
inference, we use 20 component slots by default; slots that decode to empty
supports are discarded from the final decomposable scene.

Algorithms~\ref{alg:distscene_training} and~\ref{alg:distscene_inference}
summarize the training and inference procedures of DistScene. We first
construct component-aligned scene supervision from generated assets and use
it to train Scene-Frame Generation and Object-Centric Refinement. At
inference, the environment and objects are generated jointly and each object
is subsequently refined in its local space.

\paragraph{For evaluation on the MIDI-test benchmark.}
Our model takes a complete scene image as input, including the surrounding
environment. Since the original MIDI-test inputs do not provide the full
background required by our model, we re-render the test scenes using their
provided 3D geometry, including the floor, walls, and objects. From the same
rendered scenes, we additionally render aligned object masks and depth maps
for methods that require these auxiliary inputs. All methods are evaluated
using the corresponding rendered inputs, while the original ground-truth
geometry and evaluation protocol remain unchanged.

\begin{algorithm}[t]
\caption{Training data construction and model training.}
\label{alg:distscene_training}
\begin{algorithmic}[1]
\Require Pretrained generator $\mathcal{G}_0$;
text-to-image model $\mathcal{P}_{\mathrm{img}}$;
maximum slot number $N_{\max}$
\Ensure $\mathcal{F}_{\mathrm{SS}}$,
$\mathcal{F}_{\mathrm{SLAT}}$,
$\mathcal{F}_{\mathrm{ref}}$, and
$\mathcal{F}_{\mathrm{tex}}$

\Statex \textbf{Data construction}
\For{each sampled environment and object prompt set}
    \State Generate conditioning images with $\mathcal{P}_{\mathrm{img}}$
    \State Generate the environment and canonical object assets with
    $\mathcal{G}_0$, including shape and texture SLATs
    \State Sample compatible objects, scales, and scene-space translations
    \State Place objects on valid support surfaces and reject floating or
    penetrating placements
    \State Use AABB tests followed by mesh inspection to adjust or reject
    object--object intersections
    \State Reject placements with invalid object--environment collisions
    \State Render accepted scenes from sampled viewpoints
    \State Re-encode the placed components in the shared scene frame
    \State Store rendered images, scene-frame targets, canonical object
    SLATs, and component transforms
\EndFor

\Statex \textbf{Model initialization}
\State Initialize all generation modules from $\mathcal{G}_0$
\State Freeze the pretrained backbones and optimize LoRA parameters

\Statex \textbf{Scene-frame training}
\For{each training batch}
    \State Randomly append empty object slots up to $N_{\max}$
    \State Extract scene-frame structure targets
    $\{H_i^{\mathrm{sf}}\}_{i=0}^{N}$ and their supports
    $\{S_i^{\mathrm{sf}}\}_{i=0}^{N}$
    \State Train $\mathcal{F}_{\mathrm{SS}}$ with image conditioning:
    \[
    I\longrightarrow\{H_i^{\mathrm{sf}}\}_{i=0}^{N}
    \]
    \State Decode the predicted structure latents into sparse voxel
    coordinates for all components
    \State Train $\mathcal{F}_{\mathrm{SLAT}}$ to generate scene-frame
    shape SLATs conditioned on $I$ and
    $\{S_i^{\mathrm{sf}}\}_{i=0}^{N}$
\EndFor

\Statex \textbf{Object-centric refinement training}
\For{each training scene and each object component $i$}
    \State Use the canonical object support
    $S_i^{\mathrm{can}}$ and canonical shape SLAT
    $Z_{i,\mathrm{can}}^{\mathrm{shape}}$
    \State Mark component $i$ in the scene context with a learnable
    embedding $e_i$
    \State Train $\mathcal{F}_{\mathrm{ref}}$ to recover
    $Z_{i,\mathrm{can}}^{\mathrm{shape}}$ from
    \[
    \left(
    I,\{Z_{j,\mathrm{sf}}^{\mathrm{shape}}\}_{j=0}^{N},
    S_i^{\mathrm{can}},e_i
    \right)
    \]
\EndFor

\Statex \textbf{Texture generation training}
\State Train the environment texture branch to generate
$Z_{0,\mathrm{can}}^{\mathrm{tex}}$ from the image, all scene-frame
shape SLATs, the environment support, and $e_0$
\For{each object component $i$}
    \State Train the object texture branch to generate
$Z_{i,\mathrm{can}}^{\mathrm{tex}}$ from the image, all scene-frame
shape SLATs, $S_i^{\mathrm{can}}$, $e_i$, and the local shape SLAT
\EndFor

\State Update the corresponding LoRA parameters using the SS, SLAT,
refinement, and texture flow-matching losses
\end{algorithmic}
\end{algorithm}

\begin{algorithm}[t]
\caption{Inference of DistScene.}
\label{alg:distscene_inference}
\begin{algorithmic}[1]
\Require Input image $I$; component slot number $N$;
$\mathcal{F}_{\mathrm{SS}}$, $\mathcal{F}_{\mathrm{SLAT}}$,
$\mathcal{F}_{\mathrm{ref}}$, and $\mathcal{F}_{\mathrm{tex}}$
\Ensure Textured decomposable 3D scene

\Statex \textbf{Scene-frame sparse-structure generation}

\State Initialize $N+1$ noise latents
$\{X_i^0\}_{i=0}^{N}\sim\mathcal{N}(0,I)$, where $i=0$ denotes
the environment
\State Generate sparse-structure latents jointly:
\[
\{H_i\}_{i=0}^{N}
\gets
\operatorname{FlowSample}
\left(
\mathcal{F}_{\mathrm{SS}},
\{X_i^0\}_{i=0}^{N}\mid I
\right)
\]
\State Decode sparse voxel supports
$S_i^{\mathrm{sf}}\gets\mathcal{D}_{\mathrm{SS}}(H_i)$ and discard
empty slots

\Statex \textbf{Scene-frame SLAT generation}

\State Generate scene-frame shape SLATs conditioned on the image and
decoded voxel coordinates:
\[
\{Z_i^{\mathrm{sf}}\}_{i=0}^{N}
\gets
\operatorname{FlowSample}
\left(
\mathcal{F}_{\mathrm{SLAT}},
I,\{S_i^{\mathrm{sf}}\}_{i=0}^{N}
\right)
\]
\State Decode geometry-only scene-frame components:
$M_i^{\mathrm{sf}}\gets
\mathcal{D}_{\mathrm{SLAT}}
(S_i^{\mathrm{sf}},Z_i^{\mathrm{sf}})$

\Statex \textbf{Object-centric refinement}

\For{each generated object $i=1,\ldots,N$}
    \State Estimate the scene-to-local transform $\mathcal{T}_i$ from
    the decoded mesh $M_i^{\mathrm{sf}}$, and construct
    $S_i^{\mathrm{loc}}\gets
    \operatorname{Voxelize}(\mathcal{T}_i(M_i^{\mathrm{sf}}))$
    \State Mark the scene-frame tokens of object $i$ with $e_i$ and
    generate its local shape SLAT:
    \[
    \widehat{Z}_i^{\mathrm{loc}}
    \gets
    \operatorname{FlowSample}
    \left(
    \mathcal{F}_{\mathrm{ref}},
    I,\{Z_j^{\mathrm{sf}}\}_{j=0}^{N},
    S_i^{\mathrm{loc}},e_i
    \right)
    \]
    \State Decode the refined geometry and map it back:
    \[
    \widehat{M}_i
    \gets
    \mathcal{T}_i^{-1}
    \left[
    \mathcal{D}_{\mathrm{SLAT}}
    \left(S_i^{\mathrm{loc}},
    \widehat{Z}_i^{\mathrm{loc}}\right)
    \right]
    \]
\EndFor

\Statex \textbf{Texture generation}

\State Generate and decode the environment texture:
\[
A_0\gets
\mathcal{F}_{\mathrm{tex}}
\left(
I,\{Z_j^{\mathrm{sf}}\}_{j=0}^{N},
S_0^{\mathrm{sf}},e_0
\right),
\qquad
\widehat{E}\gets
\mathcal{D}_{\mathrm{tex}}
\left(S_0^{\mathrm{sf}},Z_0^{\mathrm{sf}},A_0\right)
\]

\For{each generated object $i=1,\ldots,N$}
    \State Generate its local texture conditioned on the image,
    scene-frame context, component marker, local support, and refined
    local shape SLAT:
    \[
    A_i^{\mathrm{loc}}\gets
    \mathcal{F}_{\mathrm{tex}}
    \left(
    I,\{Z_j^{\mathrm{sf}}\}_{j=0}^{N},
    S_i^{\mathrm{loc}},
    \widehat{Z}_i^{\mathrm{loc}},e_i
    \right)
    \]
    \State Decode the textured local object and map it back:
    \[
    \widehat{O}_i
    \gets
    \mathcal{T}_i^{-1}
    \left[
    \mathcal{D}_{\mathrm{tex}}
    \left(
    S_i^{\mathrm{loc}},
    \widehat{Z}_i^{\mathrm{loc}},
    A_i^{\mathrm{loc}}
    \right)
    \right]
    \]
\EndFor

\State \Return Textured environment $\widehat{E}$ and
textured object components $\{\widehat{O}_i\}_{i=1}^{N}$
\end{algorithmic}
\end{algorithm}

\subsection{Resolution Comparison}
\label{sec:supp_resolution}

\begin{figure}[t]
  \centering
  \includegraphics[width=\linewidth,pagebox=cropbox]
  {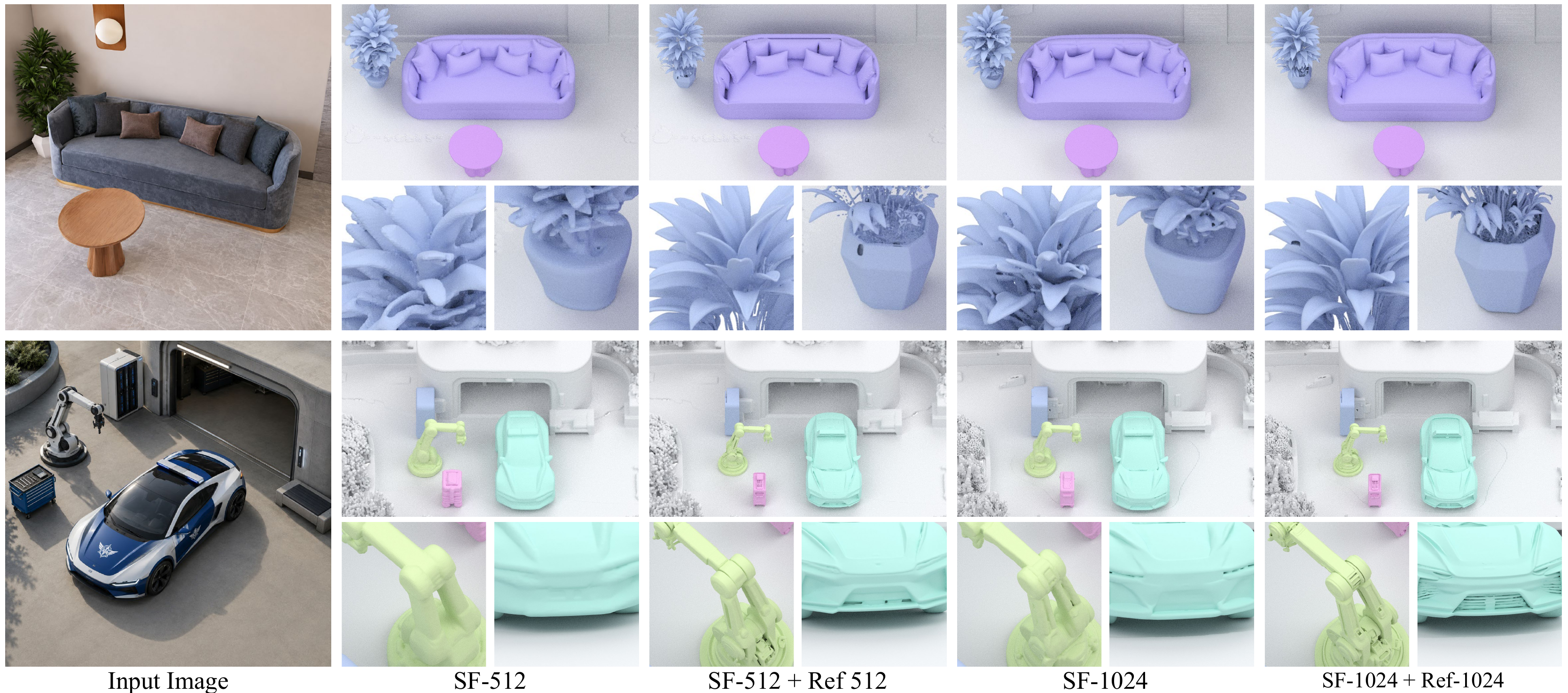}
  \caption{Qualitative comparison of generation resolution and local
  refinement. From left to right: input image, SF-512,
  SF-512 + Ref-512, SF-1024, and SF-1024 + Ref-1024.
  Each configuration is shown with a scene view and corresponding
  geometry crops. Matched views highlight differences in local detail
  while allowing comparison of the overall scene arrangement.}
  \label{fig:resolution_visual}
\end{figure}

Figure~\ref{fig:resolution_visual} compares four configurations:
SF-512, SF-512 + Ref-512, SF-1024, and SF-1024 + Ref-1024, where SF and
Ref denote scene-frame generation and object refinement, respectively.
Matched scene views and detail crops allow comparison of both the
overall arrangement and local geometry.

The examples show that similar overall arrangements can coexist with
different levels of geometric detail. In the indoor example, the
plant-leaf crops reveal differences in thin structures; in the outdoor
example, the highlighted objects and vehicle surfaces expose differences
in local shape detail. Refinement at resolution 1024 produces more
clearly resolved structures in these examples. The quantitative
quality--cost comparison is reported in
Table~\ref{tab:ablation_resolution}.

\subsection{Failure Cases and Limitations}
\label{sec:supp_failure_cases}

\begin{figure}[t]
  \centering
  \includegraphics[width=\linewidth]{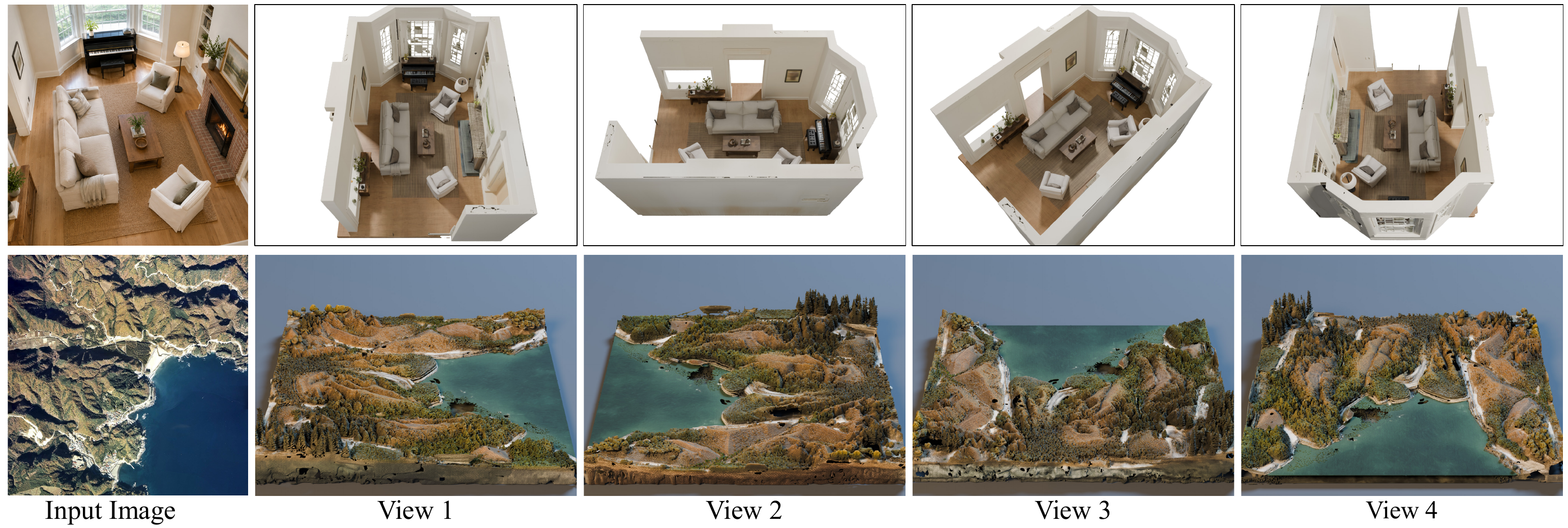}
  \caption{Representative failure cases. Indoor reconstructions may omit
  ceilings because the current environment construction relies on empty-scene
  images that rarely provide fully enclosed rooms. Outdoor reconstructions can
  exhibit holes or missing fine geometry when the scene spans a large spatial
  extent, reflecting the limited scene-frame resolution and the smaller outdoor
  training corpus.}
  \label{fig:failure_cases}
\end{figure}

Figure~\ref{fig:failure_cases} illustrates representative failure cases on
indoor and outdoor inputs. For indoor scenes, our scene construction pipeline
uses an image of an empty environment to generate the environment component
before composing object assets. In the current data-generation setup, obtaining
reliable fully enclosed empty rooms from pretrained 3D object generator is difficult; consequently, the
indoor environment assets used for training generally omit the ceiling. This
prior is reflected in some reconstructions, which may produce an open-top room
even when the input image depicts a closed interior.

For outdoor scenes, the large spatial extent of the environment distributes a
limited number of scene-frame voxels over a wide region. The resulting
representation can therefore lose fine geometric detail and may contain holes
in particularly large or complex environments. The outdoor training set is
also currently smaller than the indoor set because of limited computationally resources. Expanding the outdoor corpus and improving
large-scale scene representations are important directions for future work.

\subsection{Additional Qualitative Results}
\label{sec:supp_more_results}

\begin{figure*}[t]
    \centering
    \includegraphics[width=\textwidth]{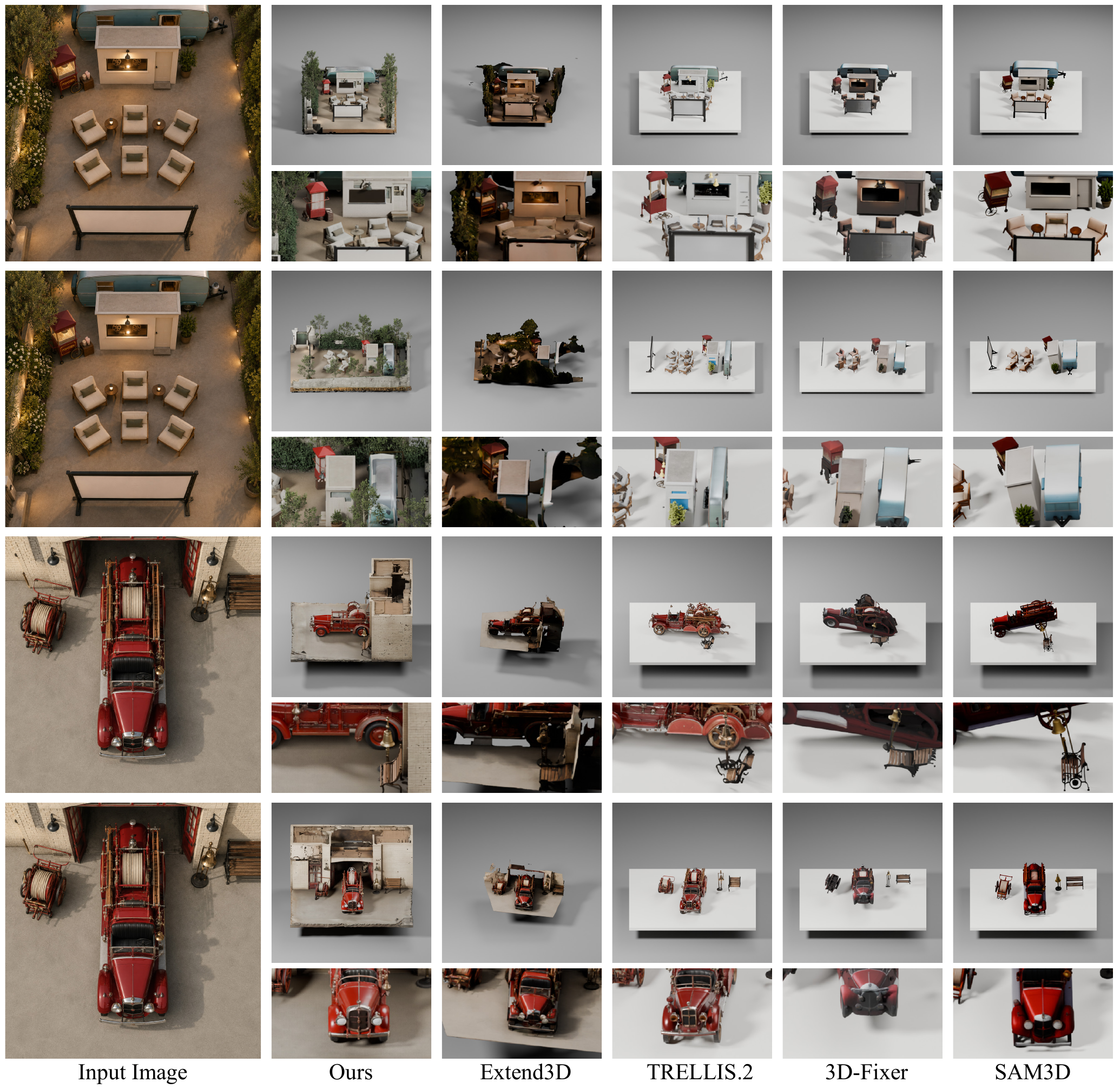}
    \caption{
        Additional qualitative comparisons between DistScene and
        Extend3D, TRELLIS.2, 3D-Fixer, and SAM3D. Each result is
        accompanied by a zoomed-in view for examining local geometric
        details. DistScene better preserves the scene layout and
        environmental context while producing more complete and detailed
        object geometry.
    }
    \label{fig:supp_more_res_1}
\end{figure*}

\begin{figure*}[t]
    \centering
    \includegraphics[width=\textwidth]{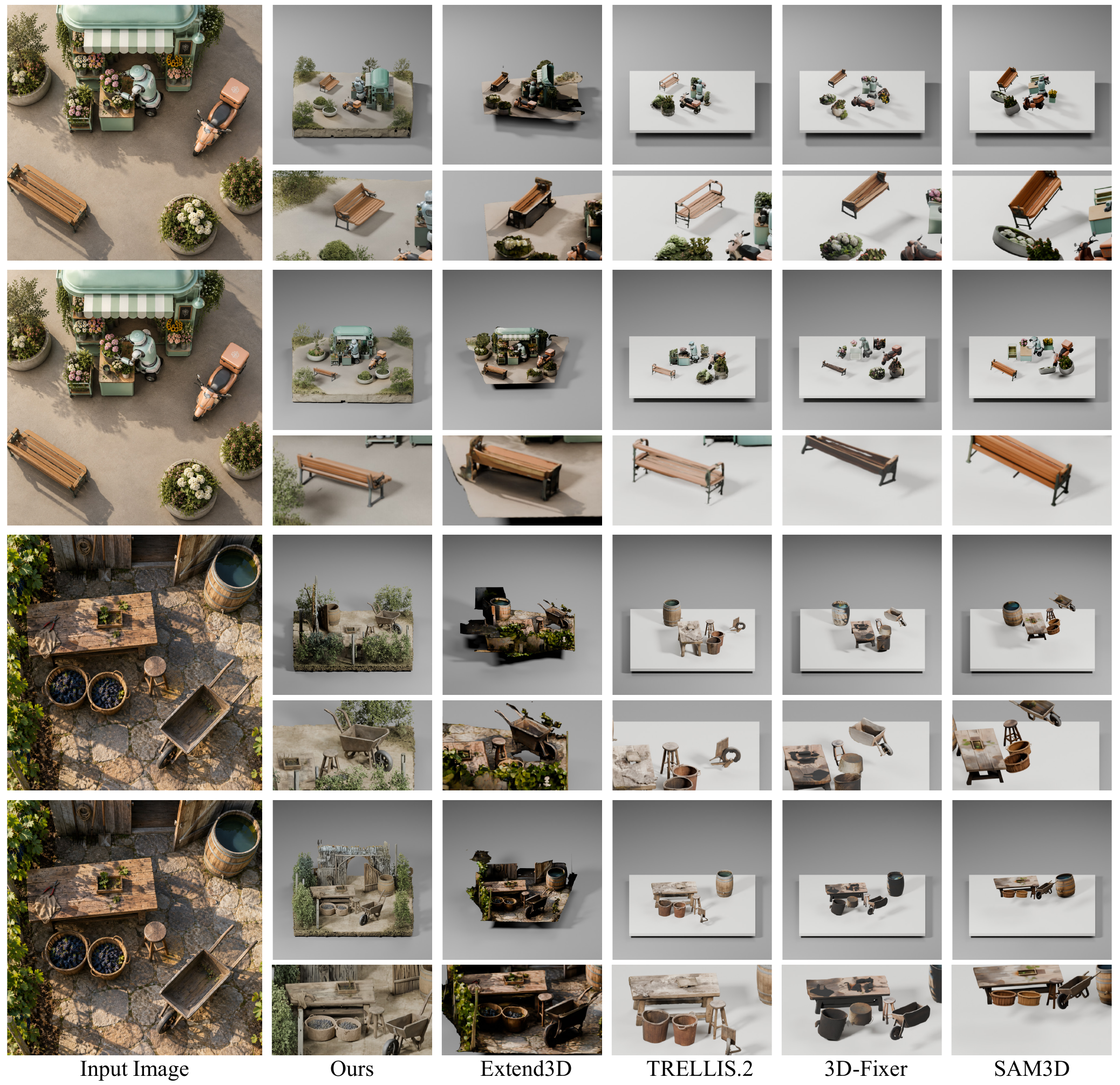}
    \caption{
        Additional qualitative comparisons between DistScene and
        Extend3D, TRELLIS.2, 3D-Fixer, and SAM3D. Zoomed-in views are
        provided for each result to highlight local geometric details.
        DistScene produces more coherent scene layouts, more complete
        environmental context, and finer object geometry.
    }
    \label{fig:supp_more_res_2}
\end{figure*}

Figures~\ref{fig:supp_more_res_1} and~\ref{fig:supp_more_res_2}
provide additional qualitative comparisons on indoor and outdoor scenes
with diverse object configurations. DistScene consistently recovers a
globally coherent layout that agrees with the input image, while generating
both the surrounding environment and individual objects with detailed local
geometry.

Extend3D optimizes a reconstructed scene using consistency between the input
image and its estimated depth. This objective can preserve the visible
appearance of the input, but provides limited constraints for heavily
occluded regions, often resulting in incomplete geometry. Moreover, Extend3D
does not explicitly represent the scene as independently addressable
components, and therefore does not directly support object-level editing.
TRELLIS.2 is an object-level 3D generator. When given a segmented
multi-object image, it can synthesize several objects and recover
recognizable local appearance. However, all objects are generated as a
single fused object-centric representation rather than as independently
indexed scene components. Without an explicit scene frame and component-wise
spatial representation, the relative positions and scales of the generated
objects are often inaccurate, while small objects receive limited spatial
support and lose fine geometric details.
3D-Fixer and SAM3D can
recover several individual objects, but their outputs may contain incomplete
geometry or missing environmental context, leading to less complete
object--environment relationships. In contrast, DistScene jointly generates
the environment and multiple independent objects in a shared scene frame,
which provides both coherent scene layout and directly editable
component-level geometry.

\end{document}